\documentclass[square]{article}

\PassOptionsToPackage{numbers}{natbib}
\usepackage[final]{neurips_2024}

\usepackage[colorlinks=true, citecolor=teal]{hyperref}   

\usepackage[utf8]{inputenc} 
\usepackage[T1]{fontenc}    
\usepackage{url}            
\usepackage{booktabs}       
\usepackage{amsfonts}       
\usepackage{nicefrac}       
\usepackage{microtype}      
\usepackage{xcolor}         
\usepackage{multirow}
\usepackage{booktabs}
\usepackage{array}
\usepackage{caption}
\usepackage{makecell}
\usepackage{amsmath}
\usepackage{amssymb}
\usepackage{colortbl}
\usepackage[table]{xcolor}
\usepackage{enumitem}
\usepackage[capitalize]{cleveref}
\crefname{section}{Sec.}{Secs.}
\Crefname{section}{Section}{Sections}
\Crefname{table}{Table}{Tables}
\crefname{table}{Tab.}{Tabs.}
\usepackage{graphicx}
\usepackage{pifont}
\usepackage{wrapfig}

\definecolor{mycolor}{RGB}{222, 236, 248}  

\title{Towards Real-World Deepfake Detection: A Diverse In-the-wild Dataset of Forgery Faces}

\author{
 Junyu Shi$^{1}$, Minghui Li$^{2}$, Junguo Zuo$^{1}$, Zhifei Yu$^{1}$, Yipeng Lin$^{1}$, \\ \textbf{Shengshan Hu}$^{1}$\thanks{Corresponding author}, \textbf{Ziqi Zhou}$^{1}$, \textbf{Yechao Zhang}$^{1}$, \textbf{Wei Wan}$^{1}$, \\ \textbf{Yinzhe Xu}$^{1}$, \textbf{Leo Yu Zhang}$^{3}$
 \\
 $^{1}$ School of Cyber Science and Engineering,
Huazhong University of Science and Technology\\
 $^{2}$ School of Software Engineering, 
Huazhong University of Science and Technology \\
 $^{3}$ School of Information and Communication Technology, Griffith University \\
 $^{4}$ School of Computer Science and Technology, 
Huazhong University of Science and Technology \\
\footnotesize{\texttt{\{hushengshan\}@hust.edu.cn}
}
}

\begin{document}

\maketitle

\begin{abstract}

Deepfakes, leveraging advanced AIGC (Artificial Intelligence-Generated Content) techniques, create hyper-realistic synthetic images and videos of human faces, posing a significant threat to the authenticity of social media. 
While this real-world threat is increasingly prevalent, existing academic evaluations and benchmarks for detecting deepfake forgery  often fall short to achieve effective application for their lack of specificity, limited deepfake diversity, restricted manipulation techniques.
To address these limitations, we introduce \textbf{RedFace} (\underline{\textbf{Re}}al-world-oriented  \underline{\textbf{D}}eepfake \underline{\textbf{Face}}),  a specialized facial deepfake dataset, comprising over 60,000 forged images and 1,000 manipulated videos derived from authentic facial features, to bridge the gap between academic evaluations and real-world necessity. 
Unlike prior benchmarks, which typically rely on academic methods to generate deepfakes, RedFace utilizes 9 commercial online platforms to integrate the latest deepfake technologies found "in the wild", effectively simulating real-world black-box scenarios.
Moreover, RedFace's deepfakes are synthesized using bespoke algorithms, allowing it to capture diverse and evolving methods used by real-world deepfake creators.
Extensive experimental results on RedFace  (including cross-domain, intra-domain, and real-world social network dissemination simulations) verify the limited practicality of existing deepfake detection schemes against real-world applications. 
We further perform a detailed analysis of the RedFace dataset, elucidating the reason of its impact on detection performance compared to conventional datasets.
Our dataset is available at: \url{https://github.com/kikyou-220/RedFace}.

\end{abstract}

\section{Introduction}

Deepfake technology, utilizing advanced AI tools like Generative Adversarial Networks (GANs)~\cite{goodfellow2020generative}, crafts convincingly false videos and images by mimicking facial features from source materials~\cite{dhariwal2021diffusion,goodfellow2014generative,ho2020denoising,wang2018high}. 
While sometimes used for entertainment or creative purposes, its misuse inflicts significant societal, political, and personal damage. 
Malicious individuals may exploit it to spread false information, distort events, or violate privacy rights, jeopardizing public trust and personal security. 
In response to the potential concerns, researchers have begun to explore novel technical approaches to aid in the identification and mitigation of the dissemination of false content. 
In parallel, recent works~\cite{li2024transferable} also explored the use of adversarial examples~\cite{song2025segment, song2025seg, wang2025breaking, wang2025advedm, advclip, zhou2023downstream, zhou2025sam2,zhou2025numbod, zhou2024securely, zhou2024darksam} as a proactive defense, embedding imperceptible perturbations into images to make them resistant to tampering or deepfake generation.
Existing deepfake detection methods~\cite{ciftci2020fakecatcher,corvi2023detection,ding2020swapped,wang2023dire} typically rely on analyzing the features of fake images, owing to the disparities in distribution and image characteristics between deepfake-generated images and genuine ones. 
Consequently, numerous datasets comprising generated fake images have been proposed to facilitate the evaluation and enhancement of detection approaches.

However, these fabricated image datasets still struggle to emulate the diverse scenarios of deepfakes in real-world settings, thereby providing inadequate and unconvincing grounds for evaluating deepfake detection methods. 
\cref{tab:datasets survey} provides a comprehensive comparison of detailed information for representative relevant datasets. 
Specifically, the current forgery image datasets exhibit the following limitations:
(1) Generic Content Focus.
Some datasets~\cite{corvi2023detection, kawar2023imagic,sha2023fake,wang2023dire,zhu2024genimage} focus on generating synthetic images of generic content, such as natural landscapes and artistic works, with limited emphasis on facial deepfake. 
(2) Incomplete Facial Coverage. Datasets specifically tailored for facial aspect suffer from issues including small scale, incomplete coverage of fabrication types, and limited diversity in fabrication methods. 
Certain datasets are dedicated to face swapping~\cite{ding2020swapped}, while others lean towards entire face synthesis~\cite{borji2022generated}, resulting in incomplete coverage of deepfake types. 
Moreover, some datasets are constructed based on single generative models~\cite{cheng2024diffusion,zhang2024genface}, such as GANs or diffusion models, imposing limitations on deepfake methods.
(3) Academic Replication Bias. The current datasets for deepfake images primarily reproduce examples from typical related research papers. However, the majority of deepfake images proliferating on the internet are generated by different users utilizing user-friendly, efficient online platforms or applications. These commercial platforms may employ entirely black-box fabrication methods with proprietary advanced technologies, resulting in stronger and more natural deepfake generation capabilities. Consequently, these datasets, relying solely on replication methods from academic papers, lack the diversity and complexity of real-world deepfakes.



\begin{figure*}[t]
    \centering
    \includegraphics[scale=0.66]{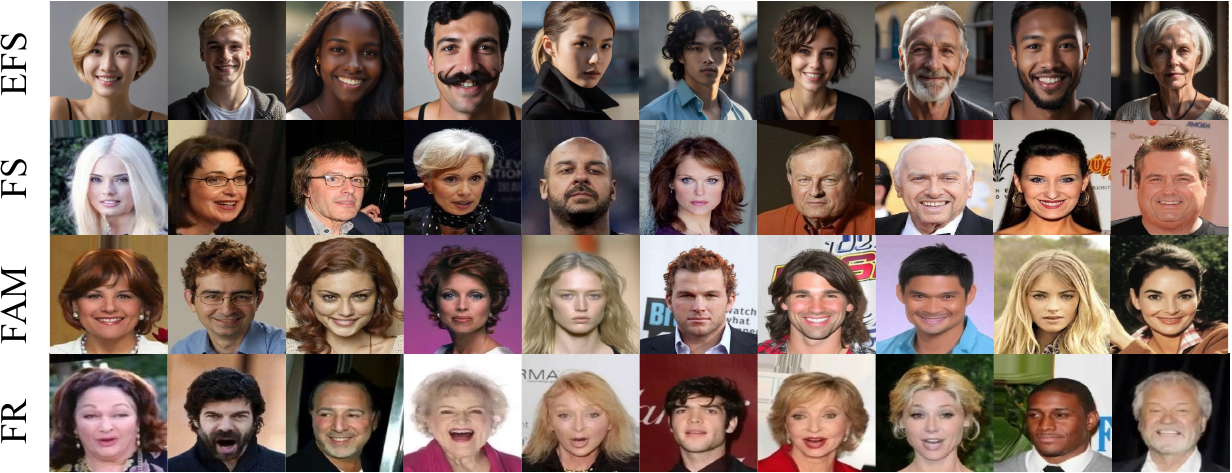}
    \caption{Visualization of images of RedFace dataset}
    \label{fig:examples}
    \vspace{-0.4cm}
\end{figure*}

To address the aforementioned issues and facilitate research into detection methods with stronger capabilities, 
we propose a novel \underline{\textbf{Re}}al-world-oriented  \underline{\textbf{D}}eepfake \underline{\textbf{Face}} dataset named \textbf{RedFace}.
Specifically, we construct our dataset based on existing literature categorizing deepfake into four main scenarios: entire face synthesis (EFS), face swapping (FS), face attribute manipulation (FAM), and face reenactment (FR)~\cite{tolosana2020deepfakes, wang2022anti}. 
We present samples from the proposed dataset in
\cref{fig:examples}.
Distinguished from other dataset construction approaches that rely on single or limited deepfake methods from academic papers, we adopt a more real-world-oriented strategy by generating deepfakes using multiple online platforms.
We believe that users in real-world scenarios are more inclined to utilize packaged services directly for generation or creation, rather than training or fine-tuning a project from scratch (due to considerations of computational resources and time cost).
Specifically, the RedFace dataset offers the following advantages: (1) Focused on deepfake scenarios, the RedFace dataset is dedicated to facial forgeries, providing more realistic and higher-resolution fake facial images compared to general datasets. (2) Comprehensive coverage of deepfake types, encompassing the four traditional categories of deepfake methods, with each image generated by these methods accompanied by comprehensive and detailed operational keyword annotations. (3) Utilization of black-box online platforms for diversified and high-quality image generation, which closely mirrors real-world deepfake scenarios.

Employing the constructed RedFace dataset, we conducted a comprehensive evaluation and comparison of current mainstream deepfake detection methods. 
Our findings reveal that many methods that perform well on specific datasets struggle to achieve comparable effectiveness when applied to image data generated in fully black-box environments, such as those of online platforms. This discrepancy underscores the vulnerabilities of these detection methods in real-world scenarios. Furthermore, we conducted comprehensive evaluations of in-domain dataset. By comparing the performance of various deepfake detection methods trained and tested on different tasks, we uncovered intriguing patterns, offering insights for future research endeavors.

\begin{table*}[ht]
\centering
\renewcommand{\arraystretch}{1.05} 
\setlength{\tabcolsep}{3.45pt} 
\caption{\textbf{Comparison of deepfake benchmark datasets:} 
The superscript ``*" indicates data regarding the video format.
\ding{51} indicates that the dataset includes data generated by the corresponding generator, while \ding{55} indicates it does not. $\bullet$ signifies support for the corresponding deepfake type, while $\circ$ signifies lack of support. \# represents the number of deepfakes or methods.}
\begin{tabular*}{\textwidth}{@{\extracolsep{\fill}} lllccccccccc}
\toprule[1.5pt]
\multirow{2}{*}{Dataset} & \multirow{2}{*}{Type} & \multirow{2}{*}{\#Deepfake} & \multicolumn{2}{c}{Generator}  & \multicolumn{4}{c}{Deepfake Type} & \multirow{2}{*}{\#Source} & \multirow{2}{*}{\#Methods}  \\ \cline{4-9}
                         &                      &     &  Diff & \multicolumn{1}{l}{GAN} & EFS     & FS     & FAM    & FR    &                             &        &                           \\ \hline
DiffForensics~\cite{wang2023dire}       &gene.    & 80k                          & \ding{51}    & \ding{55}                       & $\bullet$       & $\circ$      & $\circ$      & $\circ$  & acedamic   & 8                       \\
DMDetection~\cite{corvi2023detection}              &gene.    & 200k                         & \ding{51}    & \ding{55}                       & $\bullet$       & $\circ$      & $\circ$      & $\circ$  & acedamic    & 3                                \\
TEdBench~\cite{kawar2023imagic}       &gene.    & 0.1k                         & \ding{51}    & \ding{55}                       & $\circ$       & $\circ$      & $\bullet$      & $\circ$   & acedamic  & 1   \\
De-Fake~\cite{sha2023fake}  &gene.    & 40k                         & \ding{51}    & \ding{55}                       & $\bullet$       & $\circ$      & $\circ$      & $\circ$   & acedamic  & 2   \\
GenImage~\cite{zhu2024genimage}                &gene.     & 1300k                        & \ding{51}    & \ding{51}                       & $\bullet$       & $\circ$      & $\circ$      & $\circ$  & acedamic   & 5                                    \\ \hline
FF++~\cite{rossler2019faceforensics++}                  &facial$^{*}$         & 1k                           & \ding{55}    & \ding{51}                       & $\circ$       & $\bullet$      & $\circ$      & $\bullet$   & acedamic  & 4                                       \\
DF-TIMIT~\cite{korshunov2018deepfakes}                &facial$^{*}$       & 620                          & \ding{55}    & \ding{51}                       & $\circ$       & $\bullet$      & $\circ$      & $\circ$   & acedamic  & 1                                        \\
DF-1.0~\cite{jiang2020deeperforensics}               &facial$^{*}$        & 60k                          & \ding{55}    & \ding{51}                       & $\circ$       & $\bullet$      & $\circ$      & $\circ$  & acedamic   & 1                                              \\ \hline
SFD~\cite{ding2020swapped}                     &facial     & 420k                         & \ding{55}    & \ding{51}                       & $\circ$       & $\bullet$      & $\circ$      & $\circ$   & acedamic  & 2                                        \\
GenFace~\cite{zhang2024genface} &facial       & 500k                         & \ding{51}     &  \ding{51}                      & $\bullet$       & $\bullet$      & $\bullet$      & $\circ$    & acedamic & 12      \\
DiFF~\cite{cheng2024diffusion}                  &facial       & 500k                         & \ding{51}     &  \ding{55}                      & $\bullet$       & $\bullet$      & $\bullet$      & $\circ$   & acedamic  & 13           
\\ 
GFW~\cite{borji2022generated}                   &facial       & 15k                          & \ding{51}    & \ding{55}                       & $\bullet$       & $\circ$      & $\circ$      & $\circ$  & commercial   & 3                                    \\

\hline
RedFace \textbf{(Ours)}            &  facial       &  60k                         & \ding{51}    & \ding{51}                       & $\bullet$       & $\bullet$      & $\bullet$      & $\bullet$   & commercial  & 11                                          \\ \bottomrule[1.5pt]
\end{tabular*}
\label{tab:datasets survey}
\end{table*}
\section{Related Works}

\subsection{Deepfake Technologies}
Deepfake, a portmanteau of "deep learning" and "fake", refers to the creation of highly realistic synthetic content using deep learning techniques. Following the established categorization in existing literature~\cite{tolosana2020deepfakes, wang2022anti}, deepfakes can be primarily divided into four types.

\textbf{Entire Face Synthesis.} \textit{This involves generating virtual human faces from scratch, which, although highly realistic, do not correspond to any real individual.} Based on the underlying principles, entire face synthesis can be categorized into four main types: those based on Generative Adversarial Networks (GANs)~\cite{abdal2019image2stylegan, karras2020analyzing}, Variational Autoencoders~\cite{zhang2018stacking}, Autoregressive Models~\cite{van2016conditional}, and Diffusion Models~\cite{ho2020denoising}. Each of these techniques offers unique advantages in terms of image quality, generation speed, or detail fidelity.

\textbf{Face Swapping.} \textit{This technique replaces one person's face with another one's while maintaining the original identity's background and body.}
Early face-swapping techniques were predominantly based on traditional graphic methods to achieve facial exchange~\cite{guo2019face,nirkin2018face}. GAN-based face-swapping methods leverage adversarial training between two neural networks—the generator and the discriminator—to produce realistic face-swapped images~\cite{xu2022styleswap,yoo2023fastswap}. Moreover, by defining face-swapping as an inpainting task based on diffusion models, conditional diffusion models generate face-swapped images that balance identity replacement and attribute retention~\cite{kim2022diffface, zhao2023diffswap}.

\textbf{Face Attribute Manipulation.} \textit{This modifies specific facial attributes such as age, gender or hairstyle, while keeping other features unchanged.} Early facial attribute manipulation models modified a single attribute through data-driven training methods~\cite{shen2017learning, zhou2017genegan}. Later, multi-attribute manipulation methods emerged, addressing the challenge of disentangling attributes~\cite{gao2021high,shen2020interpreting, xu2022transeditor}.

\textbf{Face Reenactment.} \textit{This captures one person's facial expressions and maps them onto another face, enabling the target face to display the same expressions.} Methods based on 3D Morphable Models construct facial parameter models for transferring information between source and target~\cite{gao2023high, koujan2020head2head, yang2022face2face}, while others leverages self-supervised learning to achieve high-fidelity facial reenactment under specific conditions~\cite{oorloff2023robust, tripathy2020icface}.

\subsection{Deepfake Datasets}
The early definition of deepfake was confined to the generation of fabricated videos through face swapping, resulting in the introduction of numerous fake video datasets.
As shown in \cref{tab:datasets survey}, FF++~\cite{rossler2019faceforensics++} includes 1,000 real face videos sourced from YouTube, which have been manipulated using four different face swapping and face reenactment methods. Similarly, DF-TIMIT~\cite{korshunov2018deepfakes} employs the open-source software faceswap-GAN to alter real videos from the VidTIMIT database. DF-1.0~\cite{jiang2020deeperforensics} proposes an improved deepfake method called DF-VAE, which significantly enhances the quality of generated fake videos.
 
Swapped Face Detection (SFD)~\cite{ding2020swapped} is the first large-scale dataset for face swapping created using static images, which generates deepfake images by applying two face-swapping methods to source images downloaded from Google Images. 
In recent years, deepfake datasets have evolved to become more generalized, beginning to explore non-facial images.
Diffforensics~\cite{wang2023dire}, DMDetection~\cite{corvi2023detection} and De-Fake~\cite{sha2023fake} utilize 8, 3, and 2 diffusion model-based methods, respectively, to generate diverse fake images, while GenImage~\cite{zhu2024genimage} incorporates both GANs and diffusion models, resulting in a dataset of over one million images.
TEdBench~\cite{kawar2023imagic} is a small-scale dataset specifically designed for image editing, containing deepfake images that have been manipulated using their proposed methods.
Dedicated deepfake datasets for faces are relatively rare. GFW~\cite{borji2022generated} is an "in the wild" face dataset that uses three diffusion model-based online platforms to synthesize fake facial images. DiFF~\cite{cheng2024diffusion}, on the other hand, employes 13 diffusion models to create a large-scale face dataset specifically focused on diffusion models. 
Compared to DiFF, 
while GenFace~\cite{zhang2024genface} examines scenarios generated by GANs, its range of diffusion methods lacks the diversity found in DiFF.

\subsection{Deepfake Detection}
The key to deepfake detection methods lies in detecting inconsistencies introduced during the forgery process. 
%
To detect anomalies caused by AI forgery, various detection methods from different perspectives have been proposed. 
Xception~\cite{chollet2017xception} and CViT~\cite{dosovitskiy2020image} represent methods that directly employ basic backbones to train binary classifiers for learning and detecting deep forgery features. The former utilizes a CNN network, while the latter employs a Transformer architecture.
Additionally, specialized methods for detecting facial forgery, such as F3Net~\cite{qian2020thinking} and GramNet~\cite{liu2020global}, have been introduced. F3Net leverages frequency-based insights to detect realistic facial manipulations, maintaining good performance even in low-quality media. 
GramNet utilizes global image texture representations to achieve robust detection of fake images. Furthermore, specialized detection methods targeting different generators used in deepfake techniques have been proposed. PIRE~\cite{wang2023dire} is introduced specifically for detecting diffusion model forgeries, while UFD~\cite{ojha2023towards} performs real-vs-fake image classification using an untrained feature space, achieving efficient detection of fake images generated by various generative models.

\section{Dataset Design}

\begin{figure*}[t]
    \centering
    \includegraphics[scale=0.5]{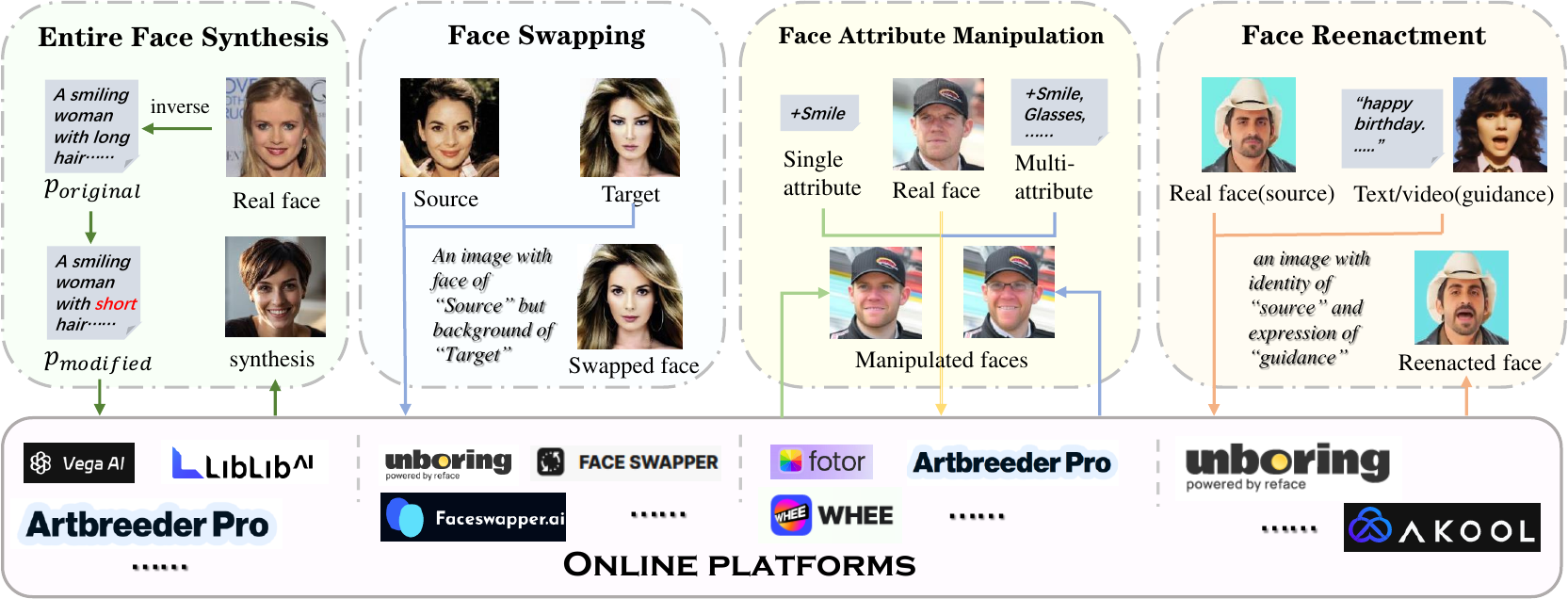}
    \caption{An overview of our benchmark scheme}
    \label{fig:pipeline}
\end{figure*}

\subsection{Authentic Face Collection}
To collect diverse authentic facial features, we leverage the CelebA~\cite{liu2018large} dataset, a widely utilized dataset for facial-related tasks, to extract an initial set of real facial images for our study. 
CelebA comprises 202,599 images of 10,177 different celebrities. Each image is annotated with 40 facial attribute labels, such as gender, hairstyle, eyeglasses, smile, and beard. Additionally, the images exhibit a variety of poses, backgrounds, lighting conditions, and facial expressions, encompassing a diverse range of facial features, making them an excellent source for deepfake operations. 
Given the variability in image quality within CelebA, with some images lacking the clarity necessary for high-fidelity deepfake manipulation, we initially assemble a meticulously curated collection of authentic facial images, referred to as dataset $I_{AUTH}$.
Specifically, we categorize the images by celebrity identity and manually select 1,023 identities with diverse distributions of attributes such as age and gender. For each identity, we chose 1 to 10 representative images that are high-resolution with  clear facial features, and are predominantly front-facing (with yaw angles within ±30 degrees and pitch angles within ±15 degrees), resulting in a total of 7411 images.

\subsection{Deepfake Generation Platform}
To obtain realistic deepfakes, we manually select the generated results to ensure a high-quality online platform where the generated effects are difficult for the human eye to discern. 
Among these platforms, some utilize GANs as their underlying architecture, while others employ diffusion models. There are also platforms with unspecified detailed architectures, yet they still produce satisfactory results. Specifically, these online platforms are as follows.
\begin{itemize}
    \item \url{https://www.vegaai.net/} Vega AI offers various generation tools, employing advanced deep learning algorithms such as diffusion models and GANs, to ensure the generated images possess high fidelity and detail clarity. We employ it for EFS.
    \item \url{www.liblib.art} Users can utilize LibLib to input various prompts, generating images with personalized characteristics, emphasizing creativity and individual expression, with specific AI techniques undisclosed. We employ it for EFS.
    \item \url{https://www.artbreeder.com/} Artbreeder is an online platform for image generation and editing that employs GANs to create and blend images. It boasts an intuitive interface and powerful image generation capabilities. We employ it for EFS and FAM.
    \item \url{https://reface.ai/unboring} The platform offers face-swapping functionality, allowing users to replace their facial features with those of another person in photos or videos, as well as creating realistic reenactment videos, with specific generation methods undisclosed. We employ it for FS and FR.
    \item \url{https://faceswapper.ai/} Faceswapper is similar to the previous one, but it can generate more realistic images. 
    \item \url{https://icons8.com/swapper} Swapper is an online tool that allows users to quickly and easily change elements within images.
    \item \url{https://www.fotor.com/} Fotor is a comprehensive online image editing tool that offers a wealth of editing features, including AI-driven editing, with specific AI methods undisclosed. We utilize it for FAM.
    \item \url{https://www.whee.com/} WHEE provides image generation and editing functionality, enabling precise attribute editing controlled by users, with specific deep methods undisclosed. We employ it for FAM.
    \item \url{https://akool.com/} AKtool is a high-quality, multi-functional image editing platform that offers efficient facial reenactment functionality, performing excellently even in low-resolution images, with specific methods undisclosed. We utilize it for FR.
\end{itemize}

\subsection{Forged Face Generation}

\begin{figure*}[t]
    \centering
    \includegraphics[scale=0.65]{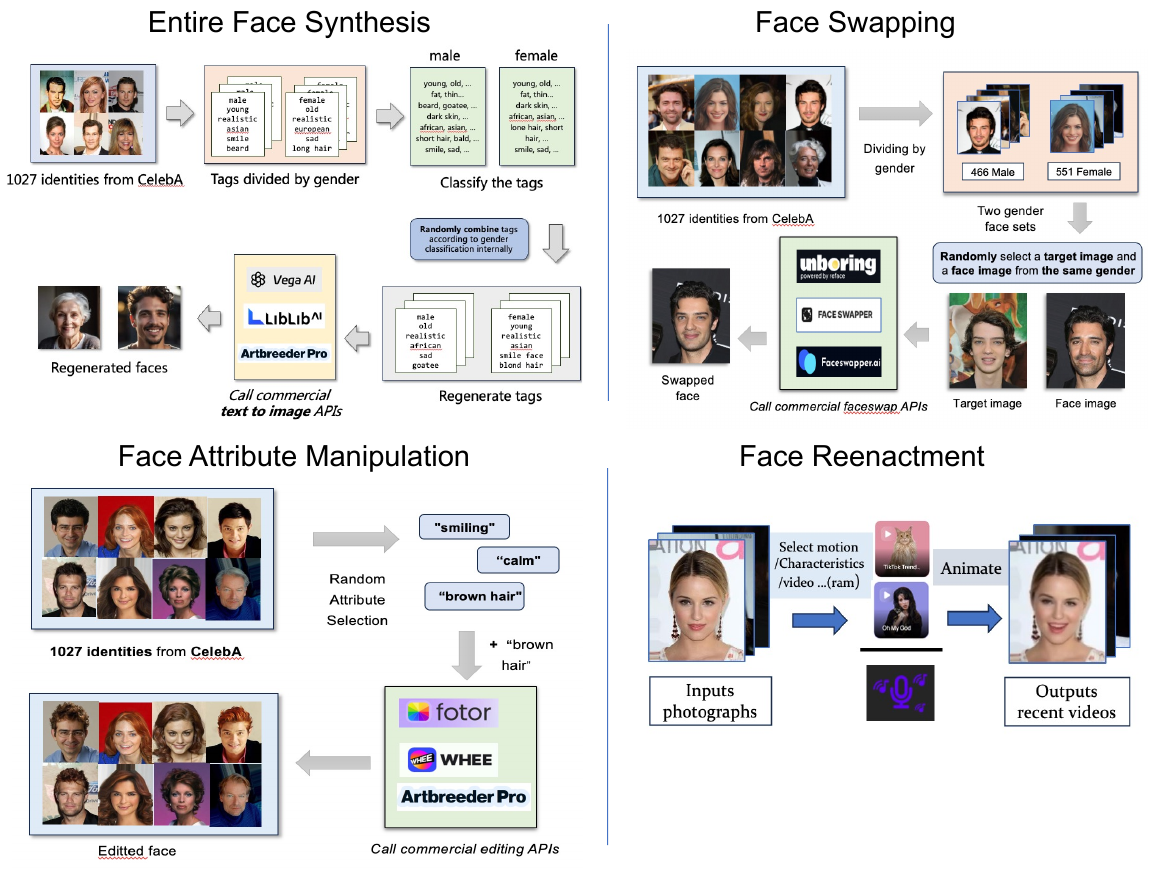}
    \caption{Process of our deepfake generation}
    \label{fig:method}
    \vspace{-0.4cm}
\end{figure*}

We categorize different types of deepfakes and design tailored strategies accordingly to utilize corresponding commercial online platforms for generating high-quality deepfake images. An overview of the process is illustrated in \cref{fig:method}.

\begin{wrapfigure}{r}{0.42\textwidth}
\vspace{-5mm}
\begin{center}
\includegraphics[clip,trim=0 0 0 0,width=0.42\textwidth]{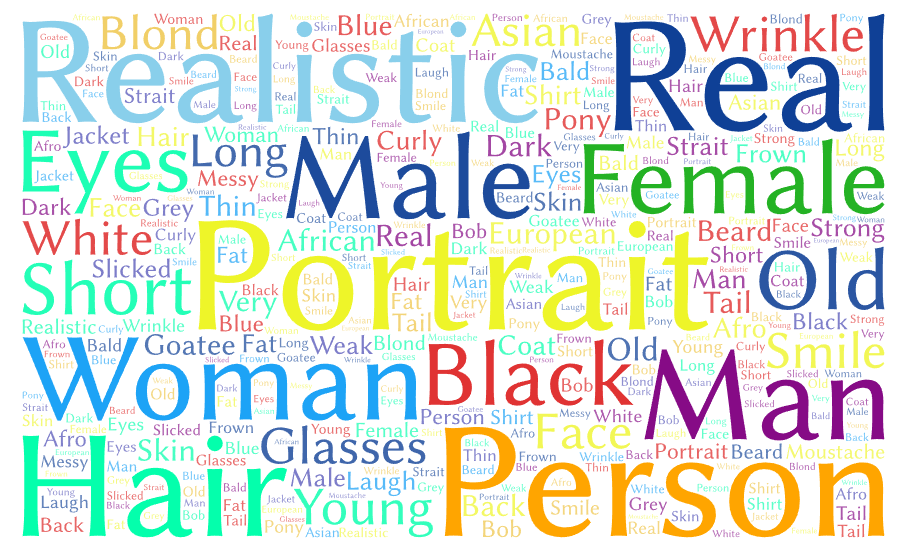}
\end{center}
\vspace{-4mm}
\caption{Word frequency distribution in face synthesis descriptive terms}
\vspace{-1mm}
\label{fig:words}
\end{wrapfigure}
 \textbf{Entire Face Synthesis.}

To generate diverse, attribute-balanced, and visually natural high-quality synthetic facial images, we implement a multi-stage prompt generation and a refinement strategy. Initially, we employ text inversion tools~\cite{promptinversion} to create initial prompts, then introduce synonyms, added contextual descriptions, and varied attribute combinations to further enhance the diversity and specificity of the prompts. 
The final word frequency distribution is depicted in \cref{fig:words}, revealing an even distribution across major attributes with a notable emphasis on realism. 
This process enriches each prompt, making them more detailed and varied. 
Subsequently, we utilize multiple facial synthesis platforms to input these refined prompts and generate facial forgery images. 
The prompts used for synthesis are annotated.

\textbf{Face Swapping.}
To achieve high-quality face swapping, we first assess face swapping platforms to understand their capabilities in handling images with significant attribute differences. Through evaluation, we find that within images of the same gender, regardless of other attributes such as age, hairstyle, and expression, the platforms demonstrate good generation capabilities. 
However, face swapping across genders often results in noticeable artifacts.
Therefore, we extract detailed attribute annotations from dataset $I_{AUTH}$ and group  the original images based on gender. 
Within different gender groups, we employ a random selection strategy to choose a source-target image pair for face swapping on online platforms through feeding the  image pair and adjusting parameters. 
The face swapped images have been undergone rigorous visual evaluation and quality control to ensure correct facial alignment, consistent lighting, and skin tone matching. Both the source and target images for face swapping are annotated.

\textbf{Face Attribute Manipulation.}
We employ a strategy of random attribute selection and multi-step editing to ensure diverse and high-quality edited images. 
The CelebA dataset, with its 40 attributes, provides a rich selection space for our modifications. 
Specifically, we randomly select  1 to 3 attributes for editing via online platforms. 
We first perform single-attribute edits on randomly chosen images. 
For instance, we select an image labeled as "no glasses" from the CelebA dataset and edit it to "wearing glasses" by platforms. 
Once single-attribute edits are completed and the image quality is satisfactory, we incrementally add other attributes for multi-attribute editing, e.g., starting with "wearing glasses", we then add "smiling" and "blonde hair". Furthermore, after each edit, we conduct rigorous quality assessments, including visual evaluation and consistency checks, to ensure the images to appear realistic and the modified attributes are accurately reflected. The target attributes modified by the editors are labeled corresponding to the source images.

\textbf{Face Reenactment.}
Source images are selected for their varied facial features, genders, and age groups, paired with guiding texts or videos depicting emotions and mouth movements from diverse contexts. 
To ensure naturalness, we match source images with guides of similar facial features, postures, and lighting conditions. 
Using platforms like Unboring and AKOOL, we input these pairs and obtain the reenacted videos as output, saving them accordingly. Simultaneously, we provide guidelines and annotations for the output alongside the source images.


\section{RedFace Benchmark}
We select several state-of-the-art deepfake detectors based on different principles for evaluation on our RedFace dataset. We include the CNN-based model Xception~\cite{chollet2017xception} and the transformer-based model CViT~\cite{wodajo2021deepfake}, as well as detectors specifically designed for facial forgery detection, such as F3Net~\cite{qian2020thinking} and GramNet~\cite{liu2020global}. 
Additionally, considering the underlying mechanics of forgery methods, we incorporate detectors like DIRE~\cite{wang2023dire}, which is tailored for diffusion model forgeries, and UFD~\cite{ojha2023towards}, a universal detector generalizable to various generative models.
In our experiments, we use accuracy (ACC) and area under the curve (AUC) as evaluation metrics to comprehensively assess the performance of deepfake detectors. For simplicity, we express them all as percentages.

\subsection{Cross-domain Evaluation}
We first evaluate the detection performance of different methods when trained on their original data domains but tested on our RedFace dataset. Specifically, these detection methods are initially tailored to specialized datasets for their respective tasks, achieving high detection performance through targeted training. 
We train each detection method to its optimal performance based on the aforementioned principles and then assess their generalization capabilities by testing these optimal models on various tasks within our RedFace dataset. 
According to their original experimental designs, 
Xception, F3NET, and UFD are trained using the FF++ dataset, while CViT and GramNet utilize the DFDC and GenImage datasets. Additionally, DIRE is trained on the DiFF dataset.
The final training results are shown in \cref{tab:training results}, demonstrating that all models achieve detection performance consistent with their respective original papers, indicating that the detection methods have been optimized. 
The evaluation results on the RedFace dataset are presented in \cref{tab:cross-domain}. 
We observe that although all detection methods achieve good performance on their respective targeted datasets, the detection performance decrease to varying degrees when these trained models are applied to a new dataset. 
This indicates that the current deepfake detection methods still have certain shortcomings when faced with new forged data domains, highlighting the need for more robust and versatile detection approaches. 

\begin{table*}[b]
\centering
\renewcommand{\arraystretch}{1.1} 
\setlength{\tabcolsep}{5pt} 
\caption{Cross-domain training results}
\begin{tabular*}{0.5\textwidth}{@{\extracolsep{\fill}}c|ccc}
\toprule[1.5pt]
Detector & Training set & ACC   & \cellcolor[HTML]{EFEFEF}AUC   \\ \hline
Xception~\cite{chollet2017xception} & FF++         & 98.78 & \cellcolor[HTML]{EFEFEF}99.87 \\ 
CViT~\cite{wodajo2021deepfake}    & DFDC         & 91.50 & \cellcolor[HTML]{EFEFEF}91.00 \\
\hline
F3Net~\cite{qian2020thinking}    & FF++         & 99.95 & \cellcolor[HTML]{EFEFEF}99.80 \\
GramNet~\cite{liu2020global}  & GenImage     & 88.14 & \cellcolor[HTML]{EFEFEF}77.02   \\
\hline
DIRE~\cite{wang2023dire}     & DiFF         & 98.95 & \cellcolor[HTML]{EFEFEF}100   \\
UFD~\cite{ojha2023towards}      & FF++         & 98.00 & \cellcolor[HTML]{EFEFEF}99.22   \\ \bottomrule[1.5pt]
\end{tabular*}
\label{tab:training results}
\end{table*}

Notably, among these methods, Xception, F3NET, and UFD achieve good results on their original training datasets. However, when faced with completely different new datasets, they exhibite a significant performance drop, with AUC values plummeting from over $90\%$ to around $50\%$. 
This performance is comparable to random guessing in a binary classification task such as deepfake detection, indicating a near-total loss of detection capability. On the other hand, GramNet, despite achieving lower AUC scores during replication, demonstrate more stability when handling new datasets, retaining some detection performance. In various subsets of tasks, DIRE demonstrates consistent stability, perhaps owing to the integration of state-of-the-art diffusion model techniques on online platforms.

Furthermore, we observe that the performance degradation of these methods is generally less pronounced on the FR subset. This could be attributed to the fact that face reenactment data (obtained by extracting frames from reenacted videos on online platforms) more closely resembles the scenarios of previous deepfake video detection tasks. Data from datasets like FF++ and DFDC include some videos sourced from publicly available media, potentially leading to some overlap in data domains. Therefore, models trained on these datasets tend to perform better on the FR subtask.

\begin{table*}[ht]
\centering
\renewcommand{\arraystretch}{1.15} 
\setlength{\tabcolsep}{5pt} 
\caption{Cross-domain evaluation}
\begin{tabular*}{\textwidth}{@{\extracolsep{\fill}}c|cccccccc}
\toprule[1.5pt]
& \multicolumn{2}{c}{EFS}               & \multicolumn{2}{c}{FS}                & \multicolumn{2}{c}{FAM}               & \multicolumn{2}{c}{FR}                \\ \cline{2-9} 
\multirow{-2}{*}{Detector} & ACC   & \cellcolor[HTML]{EFEFEF}AUC   & ACC   & \cellcolor[HTML]{EFEFEF}AUC   & ACC   & \cellcolor[HTML]{EFEFEF}AUC   & ACC   & \cellcolor[HTML]{EFEFEF}AUC   \\ \hline
Xception~\cite{chollet2017xception}                   & 49.22 & \cellcolor[HTML]{EFEFEF}50.23 & 36.70 & \cellcolor[HTML]{EFEFEF}36.47 & 58.52 & \cellcolor[HTML]{EFEFEF}58.91 & 71.13 & \cellcolor[HTML]{EFEFEF}71.14 \\
CViT~\cite{wodajo2021deepfake}                      & 51.32 & \cellcolor[HTML]{EFEFEF}53.86 & 59.55 & \cellcolor[HTML]{EFEFEF}61.07 & 50.13 & \cellcolor[HTML]{EFEFEF}49.57 & 68.43 & \cellcolor[HTML]{EFEFEF}67.34 \\ \hline
F3Net~\cite{qian2020thinking}                      & 49.40 & \cellcolor[HTML]{EFEFEF}43.62 & 51.06 & \cellcolor[HTML]{EFEFEF}53.59 & 48.88 & \cellcolor[HTML]{EFEFEF}49.88 & 57.51 & \cellcolor[HTML]{EFEFEF}58.44 \\
GramNet~\cite{liu2020global}                    & 83.30 & \cellcolor[HTML]{EFEFEF}83.30 & 45.53 & \cellcolor[HTML]{EFEFEF}53.19 & 54.85 & \cellcolor[HTML]{EFEFEF}64.78 & 66.58 & \cellcolor[HTML]{EFEFEF}52.02 \\ \hline
DIRE~\cite{wang2023dire}                       & 79.66 & \cellcolor[HTML]{EFEFEF}83.04 & 72.72 & \cellcolor[HTML]{EFEFEF}56.75 & 74.22 & \cellcolor[HTML]{EFEFEF}65.95 & 81.26 & \cellcolor[HTML]{EFEFEF}93.09 \\
UFD~\cite{ojha2023towards}                        & 44.15 & \cellcolor[HTML]{EFEFEF}10.76 & 67.95 & \cellcolor[HTML]{EFEFEF}80.86 & 59.10 & \cellcolor[HTML]{EFEFEF}71.30 & 52.00 & \cellcolor[HTML]{EFEFEF}45.92 \\ \bottomrule[1.5pt]
\end{tabular*}
\label{tab:cross-domain}
\end{table*}

\subsection{In-domain Evaluation}
Based on the literature~\cite{cheng2024diffusion, wang2023dire}, maintaining training and testing datasets within the same domain contributes to ensuring the accuracy and consistency of model evaluation. Hence, we conduct additional in-domain experiments. 
Following~\cite{cheng2024diffusion}, the RedFace dataset is divided into training, validation, and test sets, with a ratio of 8:1:1.

\begin{figure*}[t]
    \centering
    \includegraphics[scale=0.325]{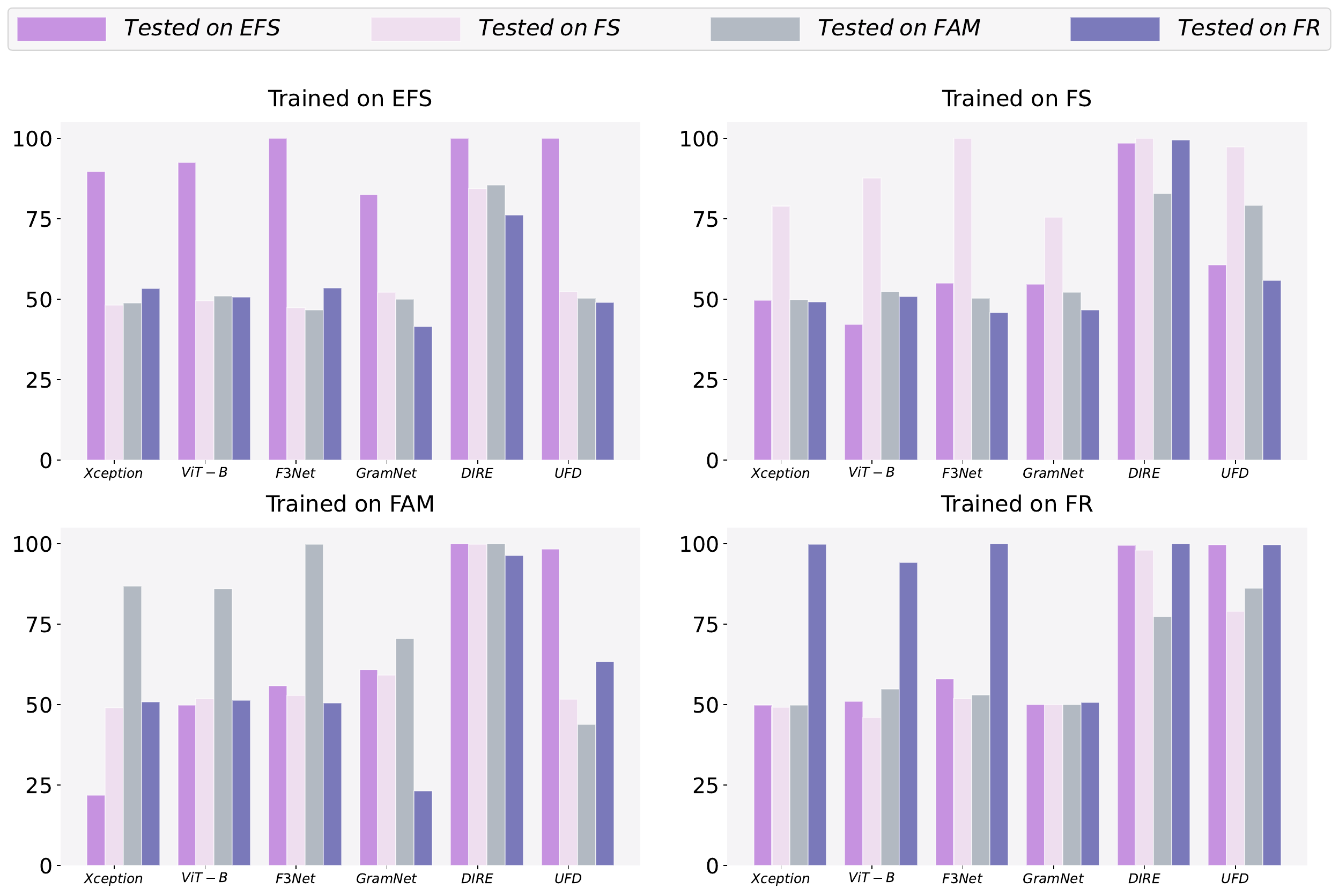}
    \caption{Cross Forgery Detection}
    \label{fig:cross forgery}
    \vspace{-0.4cm}
\end{figure*}

\textbf{Cross Forgery Detection.}
We utilize a dataset of images from a specific type of deepfake subtask as the training set to train our model, and evaluate existing detection methods on four different types of image datasets. The experimental results are shown in \cref{fig:cross forgery}. Consistent with intuition, experimental groups trained and tested on the same type of deepfake dataset exhibite better detection performance. However, in scenarios involving different types of deepfakes, detection methods designed for specific types struggle to effectively detect other types of deepfakes.
It is evident that the DIRE method outperforms other detection methods, even when trained on datasets from different tasks. Its detection accuracy and AUC remain high on datasets from other tasks, which we attribute to two factors. Firstly, online platforms likely incorporate state-of-the-art diffusion model techniques, enabling detectors specialized for diffusion models to perform well. Secondly, DIRE itself exhibits strong generalization capabilities (consistent findings in cross-domain experiments). However, DIRE's performance in cross-domain testing on FAM is slightly inferior. We attribute this to the fact that traditional attribute editing methods are mostly based on GANs rather than diffusion models. The new use of diffusion models for attribute editing needs to overcome the challenge of semantic space inversion, thus may not have been widely applied to commercial platforms. Other detection methods all exhibit remarkable declines when test on datasets from other tasks. Methods based on backbone networks like Xception and CViT, as well as F3NET, show AUCs of only around $50\%$ on other task datasets, comparable to random guessing.

\begin{table}[t]
\centering
\renewcommand{\arraystretch}{1.15} 
\caption{Evaluation (AUC) of deepfake detection schemes against degraded fake images}
\scalebox{0.68}{\begin{tabular}{c|>{\centering\arraybackslash}m{0.98cm}|cccccc|>{\centering\arraybackslash}m{0.98cm}|cccccc}
\toprule[1.5pt]
Detector  & Dataset & J70 & J30 & GB    & QN    & CR    & HR    & Dataset & J70 & J30 & GB    & QN    & CR    & HR  \\ \hline
Xception & \multirow{6}{*}{EFS} & 57.89  & 52.30  & 52.87 & 52.85 & 52.45 & 52.64 & \multirow{6}{*}{FS} & 49.72  & 49.75  & 49.63 & 49.73 & 49.72 & 49.76 \\
CViT    &                      & 51.29  & 51.32  & 50.03 & 57.41 & 50.44 & 49.31  &                     & 46.83  & 50.13  & 49.95 & 52.20 & 50.65 & 53.37  \\
F3Net    &                      & 17.82  & 19.77  & 33.02 & 49.61 & 36.28 & 29.26 &                     & 52.33  & 59.79  & 58.11 & 54.34 & 55.86 & 49.90  \\
GramNet  &                      & 84.72  & 85.96  & 88.47 & 86.11 & 81.47 & 80.59  &                     & 53.18  & 48.72  & 54.34 & 56.63 & 55.31 & 55.06  \\
DIRE     &                      & 100     & 100     & 100    & 100    & 100    & 100     &                     & 69.62       & 57.44   & 58.53    & 68.06    & 99.51    & 99.98    \\
UFD      &                      & 99.91  & 99.65  & 99.32 & 99.67 & 99.92 & 99.65 &                     & 84.16  & 77.67  & 74.71 & 80.85 & 85.40 & 50.03\\ \midrule[1.5pt]
Xception & \multirow{6}{*}{FAM} & 54.04  & 55.90  & 55.91 & 57.38 & 55.41 & 57.45 & \multirow{6}{*}{FR} & 57.08  & 55.99  & 55.83 & 57.53 & 55.24 & 57.12  \\
CViT    &                      & 55.61  & 53.73  & 48.71 & 49.39 & 53.71 & 51.93  &                     & 55.28  & 56.37  & 51.23 & 56.53 & 53.58 & 52.13  \\
F3Net    &                      & 52.59  & 55.47  & 47.36 & 55.96 & 39.34 & 51.21  &                     & 54.08  & 46.38  & 49.92 & 32.54 & 57.84 & 52.69  \\
GramNet  &                      & 56.96  & 55.63  & 55.27 & 54.35 & 53.48 & 53.52  &                     & 50.00  & 50.00  & 60.71 & 58.03 & 56.45 & 55.40 \\
DIRE     &       & 93.68     & 23.75    & 84.54    & 87.91    & 50.53    & 99.84    &                     & 99.84     & 35.68     & 95.89    & 91.58    & 99.73    & 100     \\
UFD      &                      & 45.09  & 48.29  & 35.70 & 63.85 & 51.54 & 61.33 &                     & 96.16  & 93.02  & 94.51 & 96.81 & 97.77 & 81.17 \\ \bottomrule[1.5pt]
\end{tabular}}
\label{tab:degraded}
 \vspace{-0.4cm}
\end{table}

\textbf{Degraded Forgery Detection.}
Due to the varying degrees of quality degradation that forge images undergo when circulated across different social media platforms, we simulate this process using several typical image quality reduction techniques. 
We then reassess the performance of existing detection methods on a dataset with degraded image quality. Specifically, 
We apply various data augmentation techniques to the dataset, including JPEG compression, Gaussian blur, quantization noise, color reduction, and resolution enhancement. Specifically, we set the JPEG compression parameters to 70 (J70) and 30 (J30) to simulate mild and severe compression, respectively. The Gaussian blur radius is set to 2, quantization noise parameter is set to 30, color is reduced to 64, and resolution is enhanced to the common screen resolution of 1080x1920.

We evaluate existing deepfake detection methods on the RedFace dataset, where image quality is degraded, using the optimal models trained on the same task for testing to ensure the effectiveness of the detection methods, with results shown in \cref{tab:degraded}. The experimental findings suggest that most methods experience performance degradation when detecting deepfake datasets processed by these methods, with the performance of many methods decreasing by up to $50\%$. It is noteworthy that DIRE still achieves the best performance. However, even for such a robust method, significant performance degradation occurs under severe JPEG compression ($quality=30$). This indicates that severe compression not only damages image quality but also disrupts the deepfake traces, rendering related detection methods ineffective. Furthermore, even with mild compression ($quality=70$), some methods such as Xception, CViT, and F3NET also experience significant performance declines, revealing challenges in their real-world applications.

\section{Conclusion}
This paper introduces the RedFace dataset, developed to address the limitations of existing deepfake detection datasets, which lack specificity and variety. 
Leveraging commercial online platforms, RedFace simulates real-world black-box scenarios and includes over 60k forged images and 1k manipulated videos, all with detailed annotations. 
This comprehensive dataset ensures a broad coverage of deepfake types and provides a robust framework for evaluating detection methods. 
Extensive experiments conducted on the RedFace dataset have highlighted the limitations of current detection methods, underscoring the urgent need for more powerful deepfake detection technologies.

\section*{Acknowledgements}
Shengshan Hu's work is supported by the National Natural Science Foundation of China under Grant No.62372196.
Minghui Li's work is supported by the National Natural Science Foundation of China under Grant No. 62202186. 

\bibliographystyle{iccv}
\bibliography{paper}

\end{document}